\documentclass[lettersize,journal]{IEEEtran}
\usepackage{amsmath,amsfonts}
\usepackage{algorithmic}
\usepackage{algorithm}
\usepackage{array}
\usepackage[caption=false,font=normalsize,labelfont=sf,textfont=sf]{subfig}
\usepackage{textcomp}
\usepackage{stfloats}
\usepackage{url}
\usepackage{verbatim}
\usepackage{graphicx}
\usepackage{cite}
\usepackage{multirow,amsmath}
\usepackage{color}
\begin{document}

\title{Anno-incomplete Multi-dataset Detection}

\author{Yiran Xu*, Haoxiang Zhong*, Kai Wu, Jialin Li, Yong Liu, Chengjie Wang, Shu-Tao Xia$\dagger$, Hongen Liao$\dagger$
\thanks{*: Equal Contributions}
\thanks{$\dagger$: Corresponding Authors of this paper.}
}



\IEEEpubid{}

\maketitle

\begin{abstract}
Object detectors have shown outstanding performance on various public datasets. However, annotating a new dataset for a new task is usually unavoidable in real, since 1) a single existing dataset usually does not contain all object categories needed; 2) using multiple datasets usually suffers from annotation incompletion and heterogeneous features. 
Based on the above considerations, we propose a novel problem as ‘‘Anno-incomplete Multi-dataset Detection’’, and develop an end-to-end multi-task learning architecture which can accurately detect all the object categories with multiple partially annotated datasets. Specifically, to organically integrate these datasets with incomplete annotations, we propose an attention feature interactor which helps to mine the relations among different datasets. Besides, a knowledge amalgamation training strategy is incorporated to accommodate heterogeneous features from different sources.
Extensive experiments on different object detection datasets demonstrate the effectiveness of our methods and an improvement of 2.17\%, 2.10\% in mAP can be achieved on COCO and VOC respectively.
\end{abstract}


\section{Introduction}
\label{sec:intro}

\IEEEPARstart{O}{bject} detection has attracted a lot of attention with the fast development of deep neural networks in recent years \cite{he2016deep}. Thanks to the availability of large scale public datasets \cite{lin2014microsoft,everingham2010pascal}, object detectors such as FCOS \cite{tian2022fcos} and Faster-RCNN  \cite{ren2015faster} have achieved remarkable results and are being widely applied in real world scenarios such as Face Detection\cite{yang2016wider}, Pedestrian Detection\cite{Dollar2012PAMI} and Vehicle Detection\cite{bdd100k} etc.

However, despite that more and more datasets have been compiled, the scopes of these public or private datasets are usually predefined and limited in the sense of object categories. When it comes to a new scenario, the scope of which differs from those of existing datasets, a new dataset may become in need. For example, for autonomous driving, a dataset is usually not annotated for both traffic lights and road surface markings. Even if a new dataset is tailored for the new scenario, again, the scope of the new dataset is still predefined and limited, which unfortunately falls into and contributes to this costly circle.

\begin{figure}
    \centering
    \includegraphics[width=\linewidth] {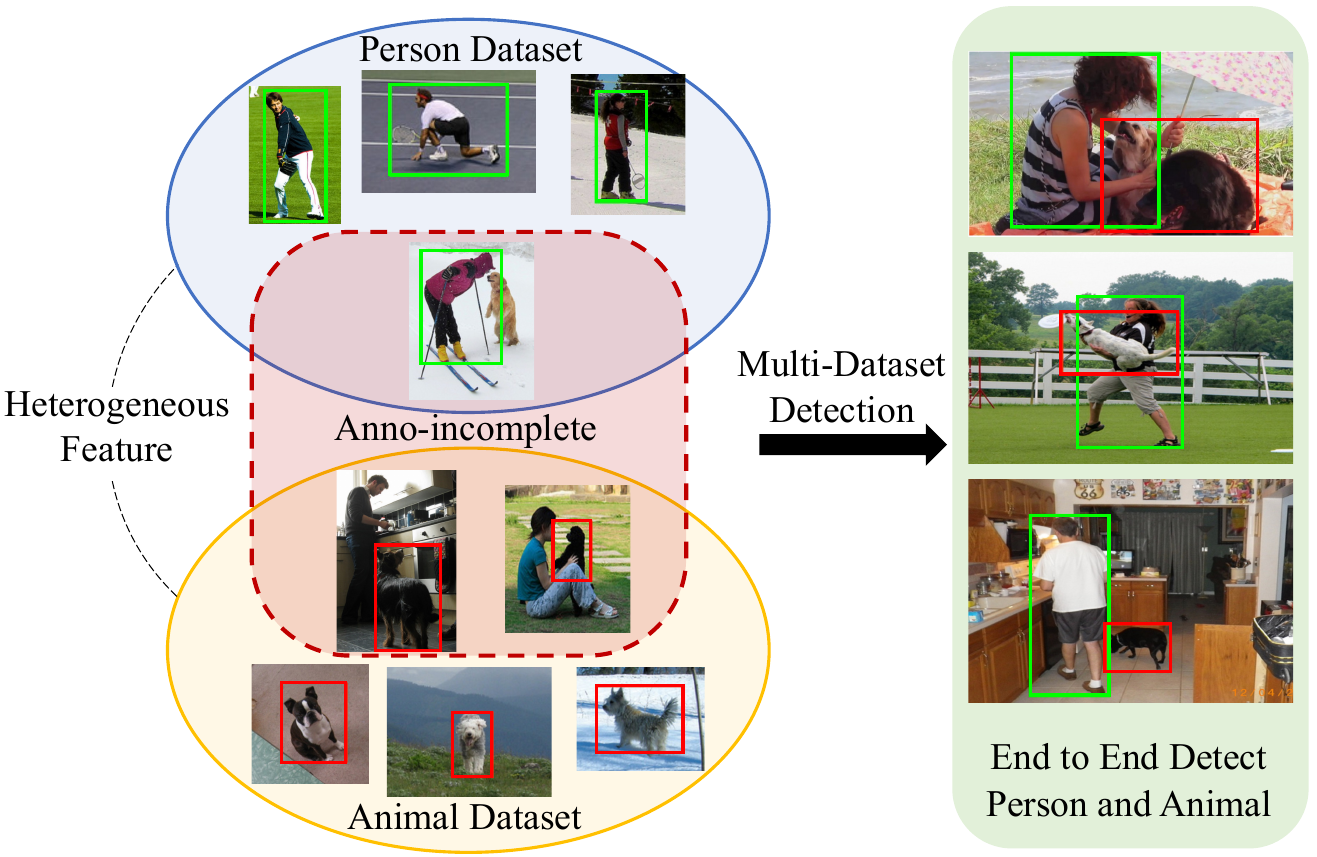}
    \caption{
    The intuition of Anno-incomplete Multi-dataset Detection.  The goal is to learn a model that can detect all the objects of multiple datasets with incomplete annotations and heterogenous features .
    Dataset A (upper-left) focuses on persons and has no annotations on dogs, in opposite to dataset B (lower-left). The target of Anno-incomplete Multi-dataset Detection (right) is to learn a model that can detect all the objects of persons and dogs from both datasets . 
    }
    \label{fig:dataset_example}
\end{figure}

Alternatively, one could work on utilizing existing datasets without manually annotating extra labels. Most of current object detectors are designed for a single dataset, and learning a model on multiple datasets of varying scopes differs from them significantly. First of all, data from different sources contains heterogeneous features, discouraging naive direct application of those detectors on the problem. Moreover, for each dataset, annotations are provided only on objects within its specific scope, but images are highly likely to contain objects outside its category coverage. It means that a dataset may contain unlabeled objects, the categories of which are not concerned in this dataset, but are in the scope of the entire problem (Fig. \ref{fig:dataset_example} gives an example). Whether making use of these unlabeled objects would help and how to effectively make use of them are also of great interest in this paper. We formulate this problem as ``Anno-incomplete Multi-dataset Detection''. To the best of our knowledge, we are the first to target on this problem. This differs from  multi-dataset detection, as we will outline in Sec.\ref{ssec:multi-dataset-detection}.

To this end, different datasets (sources) are handled in different tasks in separate branches in an integrated architecture. Recent works of multi-task learning  \cite{misra2016cross,kendall2018multi,lu2020multi, vandenhende2021multi, cui2022adaptive} usually train multiple tasks on the same set of images with full annotations, e.g. segmentation and depth estimation on the same image. In contrast, tasks are trained on different sets of images with incomplete annotations in our problem. 
Moreover, categories from different datasets could be related but with heterogeneous features. These relationships may help the detection of objects from each side. We are interested in designing effective ways to exploit the inter-dataset correlations.

\IEEEpubidadjcol

Targeting on the problem ``Anno-incomplete Multi-dataset Detection'', we propose an end-to-end cross-interactive multi-task learning framework to learn on multiple partially annotated datasets, with the goal of detecting all object categories of interest in the scope of the entire problem (also the union of the scopes of all of the datasets).
To mine the cross-branch information in the input images, upon handling different tasks in different branchs, we introduce an attention-based feature interactor via feature fusion and residual attention learning, as shown in Fig. \ref{fig:whole-net}.
Furthermore, we adopt a knowledge amalgamation training strategy to cope with heterogeneous features. With each branch supervised by each teacher model, the target
network is able to learn branch-specific information from the teachers while mining cross-branch correlations.

The main contributions of this paper are as follows:
\begin{itemize}
\item To the best of our knowledge, we are the first to formulate and target on the problem of ``Anno-incomplete Multi-dataset Detection'' for object detection.
\item We propose a branch-interactive multi-task detector where different branches handle different datasets and an attention-based feature interactor helps to mine relations implicitly among categories from different datasets.  
\item We incorporate a knowledge amalgamation training strategy to help the detector to cope with heterogeneous features, where soft pseudo labels are generated by teacher models to make use of unlabeled objects. 
\item Experiments on multiple datasets validate the effectiveness of the proposed framework, and show that our method produces an improvement of 2.17\%, 2.10\% in mAP on COCO and VOC respectively over the state-of-the-art methods in multi-dataset object detection.
\end{itemize}

\section{Related Works}
\label{sec:RelatedWorks}


\subsection{Sparse Annnotated Object Detection}
Object Detection is one of the most important research areas of computer vision. The two-stage detectors\cite{ren2015faster,cai2018cascade,chen2021high}, one-stage detectors\cite{redmon2016you,lin2017focal,tian2022fcos} and the transformer-based detectors\cite{carion2020end, dai2021updetr} have shown astonishing performance on public datasets. Efforts of many great researchers have been made targeting specific scenarios like occlusion handling\cite{kim2019bbcnet}, small objects\cite{liang2020small} and weak supervision\cite{xia2022crash}. Recently, some researchers studied the effects of sparse annotation, where the annotations of some objects in one image are missing. Efforts in this area have been made in background re-calibration loss\cite{zhang2020solving} and self-supervised learning\cite{wang2021comining}. Besides, datasets like LVIS\cite{Gupta2019lvis} are constructed in a ``federated dataset'' approach. As a result, such datasets are sparsely annotated but could be large in categories. Although ``federated dataset'' contains incomplete annotation, it is an efficient formula to annotate a large vocabulary detection dataset. Besides, all these methods develop a detector on a single dataset, which could not be used directly for detection on multiple datasets. However, our anno-incomplete multi-dataset detection aims at making use of multiple existing/public datasets to fulfill a unified new task without exhaustive data re-annotation.  

\subsection{Detection with Multiple Datasets}
To meet the need for detection on multiple datasets in the real world, three types of approaches could be used in general: direct multi-dataset detection, semi-supervised, and class incremental object detection.
\subsubsection{Direct Multi-dataset Detection}
\label{ssec:multi-dataset-detection}
Recently, some researchers began to study multi-dataset object detection, aiming at detecting objects from all these datasets with one single detector. \cite{wang2019towards} focused on handling the domain difference among datasets (e.g. face detection vs. surveillance-style detection), and managed to develop a detector to detect objects from 13 different datasets via stacks of domain attentions. 
The winner of ECCV 2020 Robust Vision Challenge \cite{zhou2021simple} tried to develop a single detector with large datasets such as COCO\cite{lin2014microsoft}, OpenImages\cite{OpenImages} and Objects365\cite{objects365}. They managed to outperform specialized detectors on each dataset by unifying the label space of these datasets and relieving the effect of confusing taxonomy of different datasets. However, they are not aware of the underlying problem of ``incomplete annotation'' in multi-dataset object detection. In this paper, we find that this problem has a great influence on the performance of the above two methods, and successfully relieve this problem implicitly by feature interaction and explicitly by an amalgamation training strategy.

\subsubsection{Semi-Supervised Object Detection}
\label{ssec:semi-supervised-object-detection}
Semi-Supervised learning (SSL) on object detection and classification has been intensively studied lately. For classification, \cite{berthelot2019mixmatch,sohn2020fixmatch,tarvainen2017mean,Berthelot2020ReMixMatch} have achieved great performance gains by pseudo-labeling, consistency and teacher-student architecture. For object detection, the pioneering works \cite{sohn2020simple} and \cite{jeong2019consistency} only focused on pseudo labeling and consistency regularization separately. To further boost the performance, \cite{tang2021humble,zhou2021instant,liu2021unbiased} combined and refined the aforementioned methods in classification. By taking one dataset as labelled and another as unlabeled, SSL can achieve multi-dataset detection with multiple models.
Our work is partly inspired by SSL but differs in many ways and attains better results. Our method is more computationally efficient as we directly utilize labels of each dataset instead of training on one dataset and semi-supervise on others. Besides, categorical relations among different datasets are explored with an AFI module. 

\subsubsection{Class Incremental Object Detection}
Incremental learning aims at learning new categories of a coming dataset without referring to the former dataset and avoiding catastrophic forgetting in this process\cite{delange2021continual}. For classification, incremental learning was achieved by sample replaying\cite{rebuffi2017icarl,lopez2017gradient}, regularization\cite{li2017learning,rannen2017encoder} or parameter isolation\cite{mallya2018packnet,serra2018overcoming}. For class incremental object detection (CIOD), sample replaying is the most prevailing approach. ILOD\cite{shmelkov2017incremental} adopted iterative supervision in the learning process by pseudo labels generated from the model of the previous iteration. Faster ILOD \cite{PENG2020109} extended this process to RPN. Recent SOTA in this area \cite{joseph2021open} focused on an open-world setting, enabling the model to detect unknown objects and learn incrementally. CIOD can detect multi-datasets by handling datasets in a streaming way. But the performance is not satisfying compared to multi-dataset detection or semi-supervised methods, as we will outline in Appendix \ref{asec:compare}.




\subsection{Knowledge Distillation and Amalgamation}
\label{sses:KD}
Knowledge distillation was first proposed by Hinton et al. \cite{hinton2015distilling}, aimed at getting a small network (student) with the ability of a larger network (teacher) by learning from the output logits or intermediate layers of the teacher.
Among them all, the most related area to ours is knowledge amalgamation \cite{shen2019amalgamating}. It aimed at utilizing homogeneous teachers in different classification tasks to distill a multi-talent student model.  Efforts in this area have been made to learn from heterogeneous networks and multi-task teachers \cite{luo2019knowledge,ye2019student}.
However, these methods mostly concerned on image classification or segmentation with no ground-truth available, which differs a lot from Anno-incomplete Multi-dataset Detection. Besides, layer-wise learning and finetuning were required, which made them hard to be applied. Instead, we design an amalgamation training strategy, where the target net is trained end-to-end and does not need any teacher at inference. 

\begin{figure*}
    \centering
    \includegraphics[width=\textwidth]{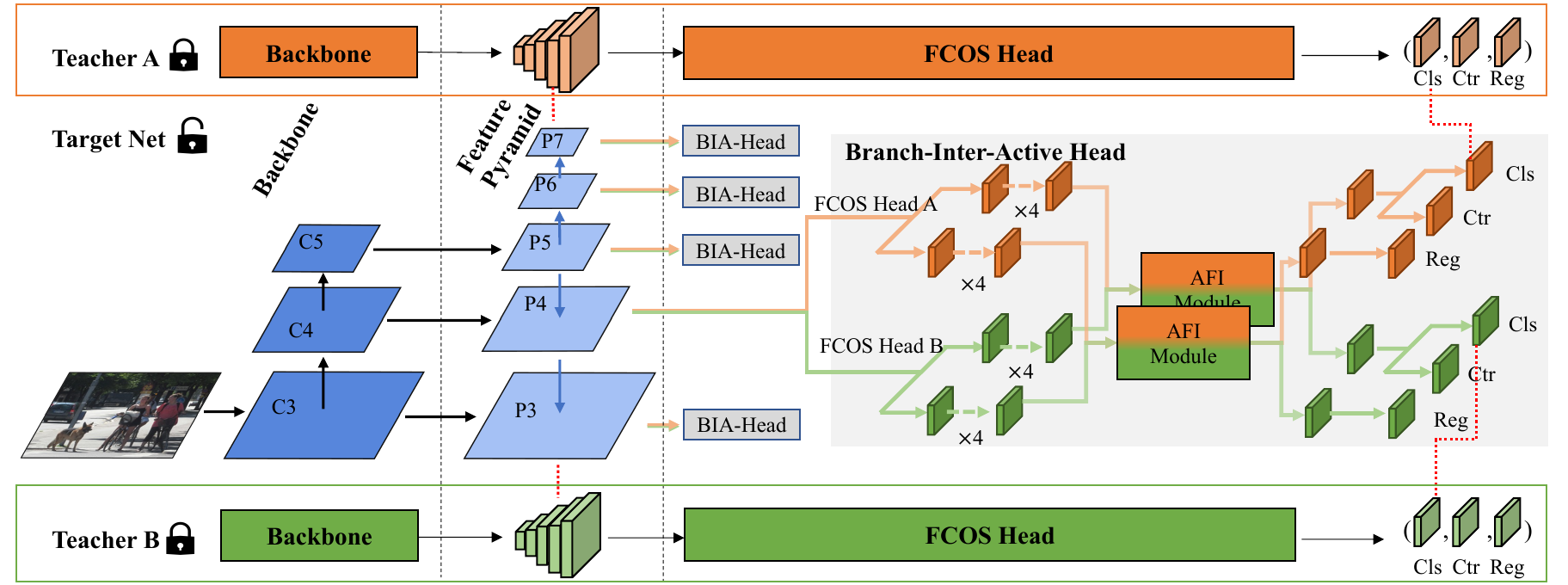}
    \caption{
    Framework of branch-interactive multi-task detector. AFI modules of each BIA head can mine the potential relations among datasets for improving unified detection performance. TeacherA and TeacherB are from amalgamation training strategy which can incorporate heterogeneous features from different sources. ``Cls'', ``Ctr'' and ``Reg'' refer to ``classification'', ``centerness'' and ``regression'' respectively.
    }
    \label{fig:whole-net}
\end{figure*}

\section{Proposed Methods}
\label{sec:ProposedMethods}

\subsection{Problem Definition} 
\label{ssec:problem}

This work aims at developing a single model for object detection using multiple partially annotated datasets, with the goal of detecting all object categories from these datasets.

Specifically, take the case of amalgamating two datasets $\mathcal{D}_{a}$, $\mathcal{D}_{b}$ as an example. The datasets $\mathcal{D}_{a}=(\mathcal{I}_{a}, \mathcal{A}_{a}, \mathcal{C}_{a}, \hat{\mathcal{C}}_{a})$ and $\mathcal{D}_{b}=(\mathcal{I}_{b}, \mathcal{A}_{b}, \mathcal{C}_{b}, \hat{\mathcal{C}}_{b})$ are from two different tasks of object detection, where $\mathcal{I}_{*}$, $\mathcal{A}_{*}$, $\mathcal{C}_{*} (*=a|b)$ are the set of 
images, the set of annotations and the set of object categories that have been annotated in the dataset respectively. $\hat{\mathcal{C}}_{*}$ is the set of all object categories that are actually contained in dataset $\mathcal{D}_{*}$, including both the annotated target categories and those that are not annotated. Meanwhile, some categories exist in both datasets but are only annotated in one of them: $\hat{\mathcal{C}}_{a} \cap \mathcal{C}_{b} \neq \emptyset$, $\mathcal{C}_{a} \cap \hat{\mathcal{C}}_{b} \neq \emptyset$ (for simplicity, assume $\mathcal{C}_{a} \cap \mathcal{C}_{b}=\emptyset$). 
Images from different sources $\mathcal{D}$ may have heterogeneous features.  
That is, objects from category set $\mathcal{C}_b$ may appear in images from $\mathcal{D}_a$ but are not annotated in $\mathcal{D}_a$, and vice versa (e.g. dogs may appear in the dataset for human detection but are not annotated). We call such kind of images as ``anno-incomplete images'', such datasets as ``anno-incomplete datasets'', and such tasks as ``anno-incomplete multi-dataset detection''. 
By leveraging both annotated and not annotated information from $\mathcal{D}_a$ and $\mathcal{D}_b$, the model is expected to detect all the object categories in $\mathcal{C}_a \cup \mathcal{C}_b$.

\subsection{Network Architecture}
\label{ssec:Network Architecture}
The proposed framework based on FCOS \cite{tian2022fcos} is shown in Fig. \ref{fig:whole-net}. The high-level design is based on multi-branch network to handle different datasets. Same as FCOS, features from different levels of the FPN are handled in turn by a shared Branch-Interactive detection Head (BIA-head). The BIA head consists of different branches of detection head (e.g. FCOS Head A/B in Fig. \ref{fig:whole-net}) as well as a feature interaction module, where each branch is responsible for the detection on one dataset. 

To mine the inter-dataset category correlations, an attention-based feature interactor merges feature maps from different branches to provide each branch with cross-branch information.
Meanwhile, to better reserve the heterogeneous features of different branches on different object detection tasks, a knowledge amalgamation training strategy is adopted during training, where each branch is also supervised by a teacher network, specifically on its object categories of interest. 
Details will be given in the following subsections.

\subsection{Attention-based Feature Interactor}
\label{ssec:Attention}
\begin{figure}
    \centering
    \includegraphics[width=\linewidth]{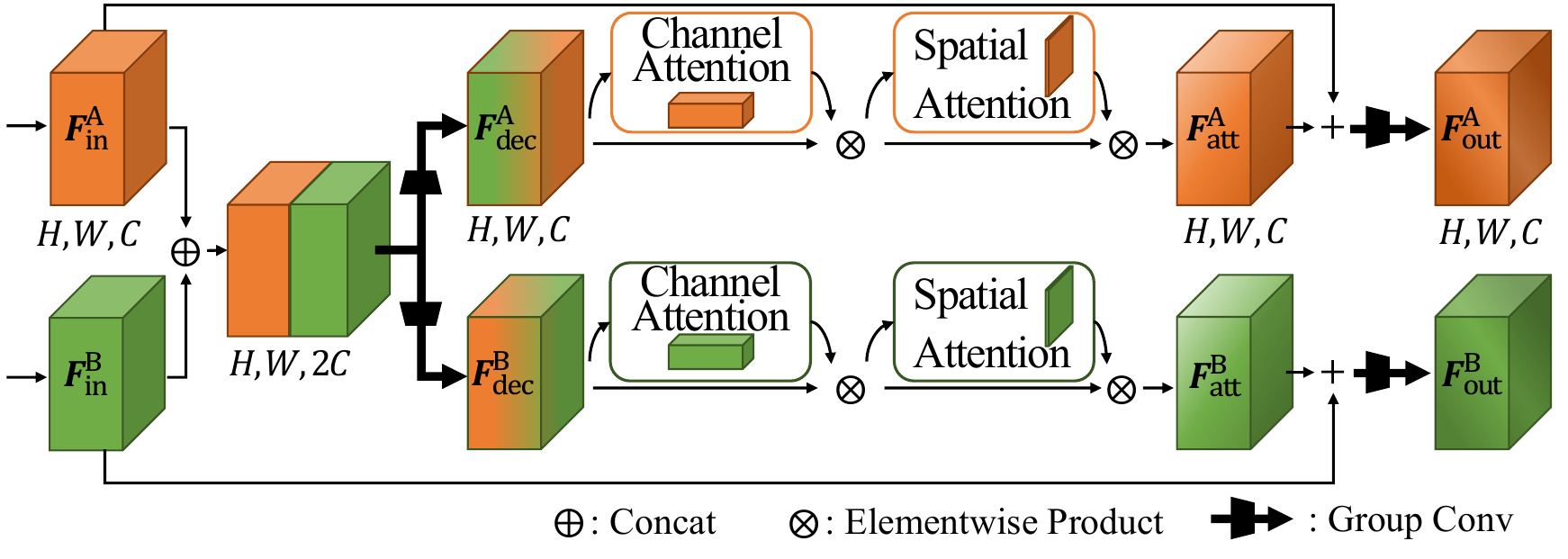}
    \caption{
    Attention-based Feature Interactor Module serves as a bridge for dataset-level interaction. The module first does feature fusion of different heads (each for a dataset). Then each head selects helpful information with an attention selection mechanism. 
    }
    \label{fig:resatt}
\end{figure}


An attention-based feature interactor (AFI) is proposed to mine potential relations among object categories from different datasets, i.e., relations among labeled objects and possibly some of the unlabeled ones within an image, so as to use the information from the unlabelled objects (usually treated as the background in conventional object detection tasks) to assist the detection of the target objects on each branch.

The structure of this module is shown in Fig. \ref{fig:resatt}. The inputs of AFI module ($\mathbf{F}_\text{in}^*, *\in\{\text{A},\text{B}\}$) are the features from different branches of the BIA head, which are decoded by separate convolutions for different object detection targets. 
$\mathbf{F}_\text{in}^\text{A}$ and $\mathbf{F}_\text{in}^\text{B}$ are firstly concatenated, and then fed into two group convolution blocks, one for each branch, to decode separate fused feature maps ($\mathbf{F}_\text{dec}^*$). On each branch, a channel attention module and a spatial attention module are applied on  $\mathbf{F}_\text{dec}^*$ to allow interaction between features from the two branches, which yields $\mathbf{F}_\text{att}^*$. Skip connections are adopted to combine $\mathbf{F}_\text{att}^*$ and $\mathbf{F}_\text{in}^*$.
Lastly, the features are refined by another group convolution block, and finally get the output features ($\mathbf{F}_\text{out}^*$) of AFI module for the subsequent classification, regression and centerness sub-tasks. 

Each group convolution block consists of a 3$\times$3 group convolutional layer of group size 16, followed by a 1$\times$1 convolutional layer, to reduce the computation load. Following the settings in FCOS \cite{tian2022fcos}, there are separate AFI modules for the regression sub-branches and the classification sub-branches between FCOS head A/B.

\subsection{Amalgamation Training Strategy}
\label{ssec:KDTraining}

Unlike most multi-task learning tasks, it is much harder for a network to learn considerably different tasks from different datasets, especially images from different datasets with heterogeneous features. For this reason, a knowledge amalgamation training strategy (KA strategy) is adopted to accommodate heterogeneous features from different sources. 
A separate model is trained beforehand for each dataset to serve as an expert in its scope of object categories. When training the branch-interactive detector (target network), the expert models (referred to as teachers in the rest of this paper) are frozen, and provide extra feature-level guidance for the target net. Besides the detection loss for images from each dataset which updates each corresponding branch and the common backbone, three other loss terms are designed for the knowledge amalgamation from the teachers on the network output as well as intermediate layers: the feature loss, the distillation loss and the pseudo loss. The amalgamation framework is shown in Fig. \ref{fig:ka-net}. Note that the teachers are only used in the training phase. During inference phase, the trained target network detects objects directly.

\subsubsection{Feature Loss}
\label{sssec:ft-loss}
\begin{figure}[t]
    \centering
    \includegraphics[width=\linewidth]{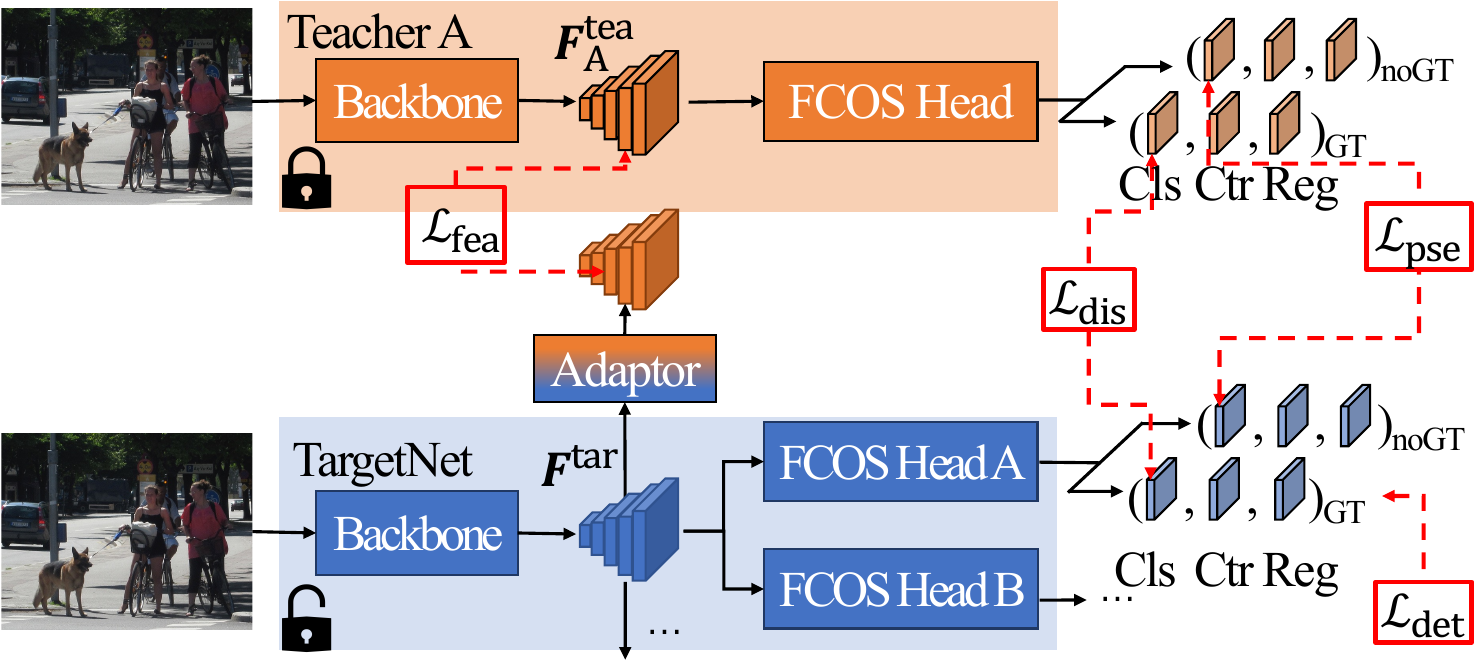}
    \caption{Framework of Amalgamation Training Strategy. Only one teacher is shown for visual simplicity. The four losses are boxed in red and also linked to the corresponding terms in red dashed lines. ``GT'' means ``ground-truth''. The locked icon at the teacher means the network is frozen. 
    }
    \label{fig:ka-net}
\end{figure}

As in  \cite{luo2019knowledge}, 
learning a backbone feature in the common space is beneficial for the target net to learn a robust representation. We adopt similar strategy and calculate feature loss at the end of backbones and FPNs (Fig. \ref{fig:ka-net}). 
1$\times$1 convolution is adopted as a feature adaptor to fuse heterogeneous teachers' feature spaces into the target net, since multiple teachers were trained independently and distilled to a single target backbone: 

\begin{equation}
\label{eq:feat_loss}
\mathcal{L}_\text{fea} =  \text{MSE}(\mathbf{F}^\text{tea} , 1\times1 \text{Conv}(\mathbf{F}^\text{tar})) ,
\end{equation}
``$\mathbf{F}$'' is the output of backbone from a teacher or target net. 

\subsubsection{Distillation Loss}
\label{sssec:kd-loss}

A distillation loss is designed at the end of the target net for the labeled data for each branch. 
Given the same input data, to align the output classification score maps of a teacher and the target net, KL divergence 
of their output classification scores is calculated to supervise the target net:
\begin{equation}
    \label{eq:kd_cls}
    \mathcal{L}_{\text{dis}} = \sum_{\{i|C^*_i>0\}} \text{KL-Div}\left(\sigma(\mathbf{C}_i^\text{tar}/T),\sigma(\mathbf{C}_i^\text{tea}/T)\right),
\end{equation}
where ``$\mathbf{C}_i$'' is the logit classification output of the $i$-th annotated anchor point/box for a branch, ``$T$'' denotes the distillation temperature and ``$\sigma$'' is the softmax function. ``$C^*_i$'' is assigned class label for $i$-th anchor point/box. This loss is only imposed on the positive samples due to severe imbalance between numbers of positive and negative samples.

\subsubsection{Pseudo Loss}
\label{sssec:pseudo-loss}

To explicitly discover and utilize those potential objects of interest, the teacher models are further exploited to generate soft pseudo labels on unlabeled images from other datasets.
The idea in pseudo labels is to extend the positive sample sets for each branch. Pseudo labels could be generated by filtering samples using logit threshold or softmax probability. The pseudo classification loss is designed as follows:
\begin{align}
    \label{eq:pseudo_cls}
    \mathcal{L}_\text{pse} &= \sum_{i\in\mathcal{S}} \text{KL-Div} \left(
    \sigma(\hat{\mathbf{C}}^{\text{tar}}_i/\hat{T}),\sigma(\hat{\mathbf{C}}^{\text{tea}}_i/\hat{T})
    \right),\\
    \label{eq:pseudo_filter}
    \mathcal{S} &=\{i|\max\sigma(\hat{\mathbf{C}}^\text{tea}_i)>\theta\},
\end{align}
where ``$\hat{\mathbf{C}_i}$'' is the logit output of the $i$-th anchor point/box for the images from other datasets (without annotations on concerned objects of the current branch). ``$\theta$'' is a threshold parameter, which filters using softmax probability.  ``$\hat{T}$'' denotes soft pseudo label temperature which control the flatness or sharpness of the softmax output distribution.  



\subsubsection{Total Loss}
\label{sssec:total_loss}
Finally, the loss function for the target net is:
\begin{gather}
    \label{eq:total_loss}
    \mathcal{L} = \sum_k (\omega_{\text{det}}\mathcal{L}_\text{det}^k +
    \omega_\text{fea}\mathcal{L}_\text{fea}^k +
    \omega_{\text{dis}}\mathcal{L}_{\text{dis}}^k +
    \omega_{\text{pse}}\mathcal{L}_{\text{pse}}^k),
\end{gather}
where ``$k$'' is the index for the branches, and ``$\mathcal{L}_\text{det}$'' is the original detection loss for each branch on the classification, regression and centerness output with the ground-truth of the corresponding dataset (more details on $\mathcal{L}_\text{det}$ are in \cite{tian2022fcos}). ``$\omega$''s are weighting coefficients.  

\subsection{Training and Evaluation Pipeline}
\label{ssec:train_eval_pipeline}

\begin{figure*}
    \centering
    \includegraphics[width=0.98\linewidth]{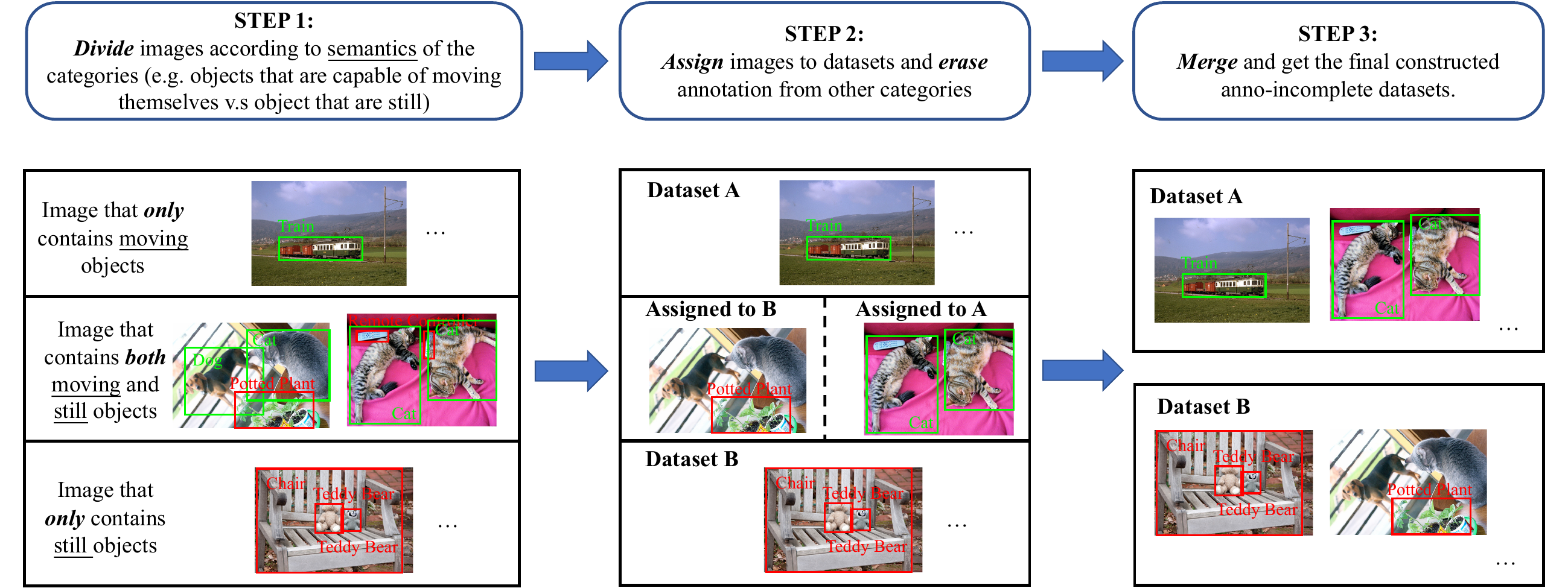}
    \caption{Pipeline of constructing anno-incomplete datasets from a given public dataset as described in Sec.\ref{ssec:Dataset}. Given the grouping of categories of the public dataset, the whole process consists of 3 steps: divide, assign and erase, and finally merge.}
    \label{fig:dataset_construction}
\end{figure*}

\subsubsection{Training Pipeline}
The algorithm proposed is a multi-branch network structure. To deal with different data sources, many multi-branch networks such as MDNet \cite{nam2016learning} adopt an iterative strategy to update each branch in turn sequentially. Alternatively, we can update all branches simultaneously. In specific, we adopted the following steps for our model training:

\begin{itemize}
    \item The specific teacher models are trained beforehand for KA strategy.
    \item Each training batch consists of the same number of samples from different datasets. With an additional dataset identifier, each detection head branch can identify labeled samples from the corresponding dataset and unlabeled samples from other datasets.
    \item After extracting features through the backbone network and FPN, the teacher network is incorporated to calculate the feature loss $\mathcal{L}_\mathrm{fea}$ for all input samples.
    \item After different detection heads decode and adjust the backbone network features through their convolution layers, the output features undergo feature interaction and fusion through the AFI module.
    \item When performing classification, regression, and centerness prediction, the output results are distinguished according to the dataset identifier. Calculation of $\mathcal{L}_\mathrm{det}$ and $\mathcal{L}_\mathrm{dis}$ are performed on the output of the labeled samples and $\mathcal{L}_\mathrm{pse}$ for the unlabeled samples.
    \item Finally, calculate the total training loss function according to  \eqref{eq:total_loss} and use it for gradient descent and parameter update of the model.
\end{itemize}

\subsubsection{Evaluation Pipeline}
During inference, the target net no longer needs the teacher networks. However, since the AFI module performs feature interaction between different branches, each branch of the target net still cannot make predictions individually. The specific process is as follows:

\begin{itemize}
    \item First, feature extraction is performed by the backbone network and FPN on the test data.
    \item Then, the features are simultaneously input to each branch of the BIA head and adjusted by their convolutions.
    \item The decoded and adjusted features of each branch undergo feature interaction and fusion through the AFI module.
    \item Each detection head branch uses the fused features to complete the subsequent classification, regression, and centerness prediction.
    \item After post-processing methods of the original detection model such as NMS, the predicted scores of each branch are turned into bounding box results. We simply merge the predicted bounding boxes as the final prediction of the target network.
\end{itemize}

\begin{table} [t]
\centering
\caption{Number of categories, images and annotations of constructed datasets from MS-COCO and PASCAL-VOC, as well as original annotations and erased percentage.
} 
\label{tb:coco_voc_partition}
\begin{tabular}{cccccc}
\hline
Dataset & \#Cat & \#Img  & \#Anno  & \#$\text{Anno}_\text{ori}$ & \%Erased\\ \hline
COCO\_A & 19         & 57893 & 271444 & 421243         & 35.56\%      \\
COCO\_B & 61         & 59373 & 293851 & 438758         & 33.03\%      \\
VOC\_A  & 7          & 8268  & 17629  & 24652          & 28.49\%      \\
VOC\_B  & 13         & 8283  & 18576  & 22571          & 17.70\%      \\ \hline
\end{tabular}

\end{table}

\begin{table*}[t]
\centering
\caption{ 
Comparison of mAP for detection methods on the split and partially annotated COCO\_A and COCO\_B. MTB is a common multi-branch network with each branch for a dataset.
``$\text{Ours}_{\leftarrow \text{R50}}$'' and ``$\text{Ours}_{\leftarrow \text{R101}}$'' means the teachers of amalgamation training strategy of our method use the backbone of Resnet-50 and Resnet-101 seperately . All target models and compared models use the Resnet-50 backbone. The best results are denoted in bold.
%
} 
\label{tb:overall}
\begin{tabular}{c|ccc|ccc|ccc|cc}
\hline
\multirow{2}{*}{Model}  &
  \multicolumn{3}{c|}{COCO} &
  \multicolumn{3}{c|}{COCO\_A} &
  \multicolumn{3}{c|}{COCO\_B} &
  \multirow{2}{*}{Params} &
  \multirow{2}{*}{FLOPS} \\ \cline{2-10}
                   & mAP   & $\text{mAP}_{50}$ & $\text{mAP}_{75}$ & mAP   & $\text{mAP}_{50}$ & $\text{mAP}_{75}$ & mAP   & $\text{mAP}_{50}$ & $\text{mAP}_{75}$ &       &        \\ \hline
MTB                & 36.18 & 54.53 & 38.85 & 50.79 & 73.32 & 55.05 & 31.63 & 48.68 & 33.80 & 36.98 M & 256.09 G \\
\hline
Single             & 36.86 & 55.15 & 39.69 & 51.12 & 73.08 & 55.53 & 32.33 & 49.43 & 34.62 & 64.30 M & 366.76 G \\
Brutal             & 35.45 & 53.17 & 38.08 & 49.98 & 72.51 & 54.34 & 30.92 & 47.15 & 33.01 & 32.24 M & 170.17 G \\
PSEUDO             & 37.38 & 55.50 & 40.34 & 51.62 & 73.72 & 55.95 & 32.96 & 49.82 & 35.44 & 32.24 M & 170.17 G \\ 
UODB               & 36.02 & 54.54 & 38.50 & 50.87 & 73.06 & 54.80 & 31.39 & 48.78 & 33.43 & 49.61 M & 256.47 G \\
SMD                & 35.64 & 53.49 & 38.42 & 50.23 & 72.72 & 54.92 & 31.07 & 47.46 & 33.32 & 32.24 M & 170.17 G \\
UBT                & 37.00 & 55.42 & 39.76 & 51.86 & 73.49 & 56.30 & 32.37 & 49.79 & 34.61 & 64.30 M & 366.76 G \\ \hline
$\text{Ours}_{\leftarrow \text{R50}}$      & 37.73 & 56.35 & 40.99 & 52.03 & 74.42 & 56.82 & 33.27 & 50.72 & 36.06 & 38.12 M & 273.81 G \\
$\text{Ours}_{\leftarrow \text{R101}}$     &\textbf{38.35} & \textbf{56.84} & \textbf{41.43} & \textbf{52.78} & \textbf{74.58} & \textbf{57.18} & \textbf{33.86} & \textbf{51.35} & \textbf{36.54} & 38.12 M & 273.81 G \\ \hline
\end{tabular}

\end{table*}

\section{Experiments on Anno-incomplete Multi-dataset Detection}
\label{sec:Experiments}

\subsection{Dataset}
\label{ssec:Dataset}

To simulate the situation on multiple related but scope-isolated datasets
(see ``Problem Definition'' in Sec. \ref{ssec:problem}), we adopt a simple method to construct anno-incomplete datasets from a given public dataset. MS-COCO \cite{lin2014microsoft} and PASCAL-VOC \cite{everingham2010pascal} are 
chosen in our experiments. Pipeline of the construction process is shown in Fig. \ref{fig:dataset_construction}.
In specific, we simply divide the categories 
into two groups according to the semantics. 
COCO categories are divided into moving objects (persons, animals and vehicles) and still objects, and VOC categories are divided into living objects (persons and animals) and non-living objects.
The images that contain annotations from both groups are randomly assigned to one of the two groups so that the total amount of images of the two groups is as balanced as possible. For each image, after the assignment into a certain group, annotations of objects from the scope of the other group is erased.

Details of the constructed subsets from training set MS-COCO 2017 and train-val set from PASCAL-VOC 2007 plus 2012 are shown in Table \ref{tb:coco_voc_partition}. Over 30\% of original annotations in COCO are erased and around 20\% for VOC (see ``\%Erased'' in Table \ref{tb:coco_voc_partition}).  

\subsection{Baselines}
\label{ssec:Baselines}
In this section, we introduce several baseline methods and SOTA methods (Table \ref{tb:overall}\&\ref{tb:overall_voc}).

``MTB'' refers to a simple multi-branch network with a shared backbone and different branches of detection heads, without feature interactor and amalgamation training strategy.
``Single'' is where each dataset has its own detector. ``Brutal'' is to simply concatenate two datasets into one and train a single-branch detector. 
``Pseudo'' is to use the ``Single'' models to predict bounding boxes on other datasets (single model for COCO\_A to infer on COCO\_B). We merge the predictions and annotations on all datasets into one, and finally train a single-branch detector. 

Besides, we also compare our method with SOTAs of multi-dataset detection ``UODB'' \cite{wang2019towards} and ``SMD'' \cite{zhou2021simple}. The SOTA for  semi-supervised detection ``UBT'' \cite{tang2021humble} can also be applied to anno-incomplete detection by taking one dataset as labeled set and others as unlabeled. 

\subsection{Experiment Settings}
\label{ssec:Settings}
\subsubsection{Training Details}
For all the experiments, our detector and compared methods are all based on the  detector FCOS \cite{tian2022fcos} with a backbone of Resnet50 \cite{he2016deep}. We use split subsets from COCO and VOC in Table \ref{tb:coco_voc_partition} as training sets. 
Details on training parameters for ours and compared methods are given in the appendices.


\subsubsection{Evaluation Details}
For COCO, all the models are evaluated on the standard validation set of MS-COCO 2017 \cite{lin2014microsoft}, where the dataset is denoted as ``COCO''. Besides, we also report results on the splits of COCO denoted as ``COCO\_A'' and ``COCO\_B''. For those that are not able to predict all categories of objects in COCO with one model (e.g ``Single''), we collect and accumulate the predictions of each split. For each split, only the categories from it are considered during evaluation. For VOC, we first convert the annotations for test set of PASCAL-VOC 2007 into COCO format, and evaluate them in COCO protocols as stated.  ``mAP'' metric \cite{lin2014microsoft} is adopted for evaluation.

\begin{table}[t]
\centering
\caption{  Comparison of mAP on VOC in COCO evaluation protocol. The best results  are denoted in bold.
}
\label{tb:overall_voc}
\begin{tabular}{c|ccc|cc}
\hline
\multirow{2}{*}{Model} & \multicolumn{3}{c|}{VOC} & VOC\_A & VOC\_B \\ \cline{2-6} 
                       & mAP    & $\text{mAP}_{50}$  & $\text{mAP}_{75}$  & mAP    & mAP    \\ \hline
MTB                    & 45.64  & 71.26  & 49.40  & 50.93  & 42.79  \\
\hline
Single                 & 45.06  & 71.76  & 48.49  & 50.96  & 41.88  \\
Brutal                 & 42.98  & 69.01  & 45.93  & 50.64  & 38.94  \\
PSEUDO                 & 42.07  & 69.39  & 44.23  & 46.35  & 39.76  \\ 
UODB                   & 44.65  & 70.63  & 48.16  & 49.86  & 41.85   \\
SMD                    & 43.14  & 68.84  & 45.93  & 50.70  & 39.12  \\
UBT                   &  46.34  & 72.87  & 49.98 & 52.26  & 43.16  \\ \hline
$\text{Ours}_{\leftarrow \text{R50}}$         & 47.07  & 72.73  & 50.82  & 52.52  & 44.13  \\
$\text{Ours}_{\leftarrow \text{R101}}$         & \textbf{47.74}   &  	\textbf{73.07}   & \textbf{51.72}  & \textbf{53.54}   & \textbf{44.62}  \\ \hline
\end{tabular}
\end{table}

\subsection{Overall Performance}
\label{ssec:OverallResults}

In this section, we compare the proposed method with the introduced baselines and SOTAs. Results on COCO and VOC are given in Table \ref{tb:overall}\&\ref{tb:overall_voc}.


Among the baselines, ``Single'' and ``MTB'' achieve the most stable results while ``Brutal'' suffers from a huge drop in mAP. While ``Pseudo'' behaves well on COCO, it is not the same case in VOC. Kindly refer to the appendices for more analysis on the behaviors of ``Brutal'' and ``Pseudo''.

Although ``UODB'' was able to perform well on its original multi-dataset settings, when it comes to anno-incomplete, the performance is not as impressive as it was.
The more recent SOTA ``SMD'' is originally designed for detection on COCO, Objects365 
and OpenImages 
by merging and unifying labels, where there are severe conflicts and confusions in label space. However, for our setting, there is no labels to merge. This almost degrades ``SMD'' to ``Brutal'' and that is why there is not much difference in accuracy between these two methods.
In general, ``UBT'' performs the best in anno-incomplete multi-dataset detection. It generates pseudo labels and makes use of other datasets besides the current one. However, ``UBT'' generates multiple models, which has heavy parameters and high computational complexity. 


Our method achieves a great improvement of 2.17\% and 2.10\% in mAP over MTB on COCO and VOC respectively, and outperforms all the others no matter if we use teacher models with a deeper backbone (Resnet-101) or not. 

\begin{figure}[t]
    \centering
    \includegraphics[width=\linewidth]{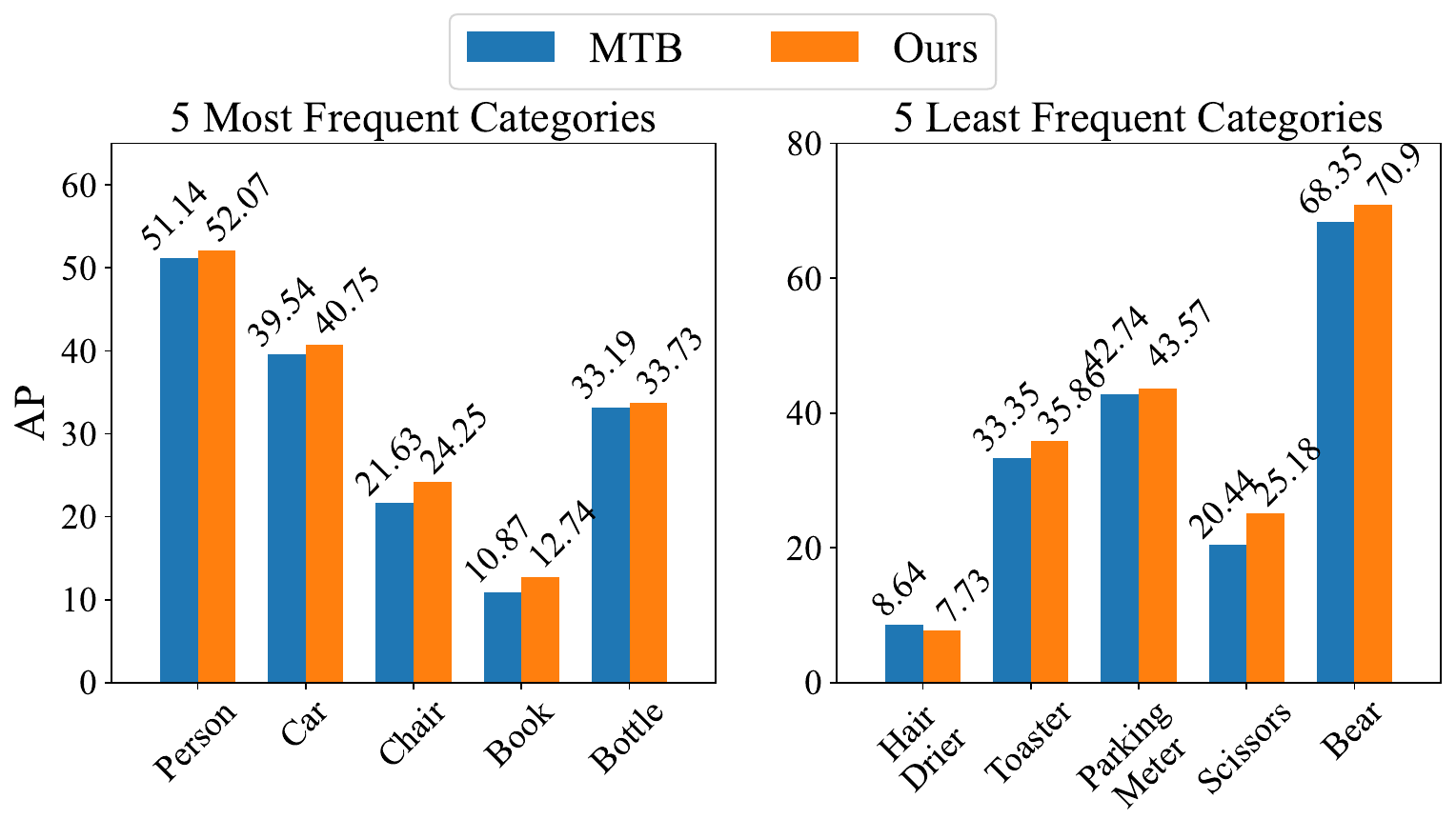}
    \caption{Comparison on AP results of the most frequent categories (left) or least frequent categories (right)  in COCO. }
    \label{fig:big5}
\end{figure}
\begin{figure}[t]
    \centering
    \includegraphics[width=\linewidth]{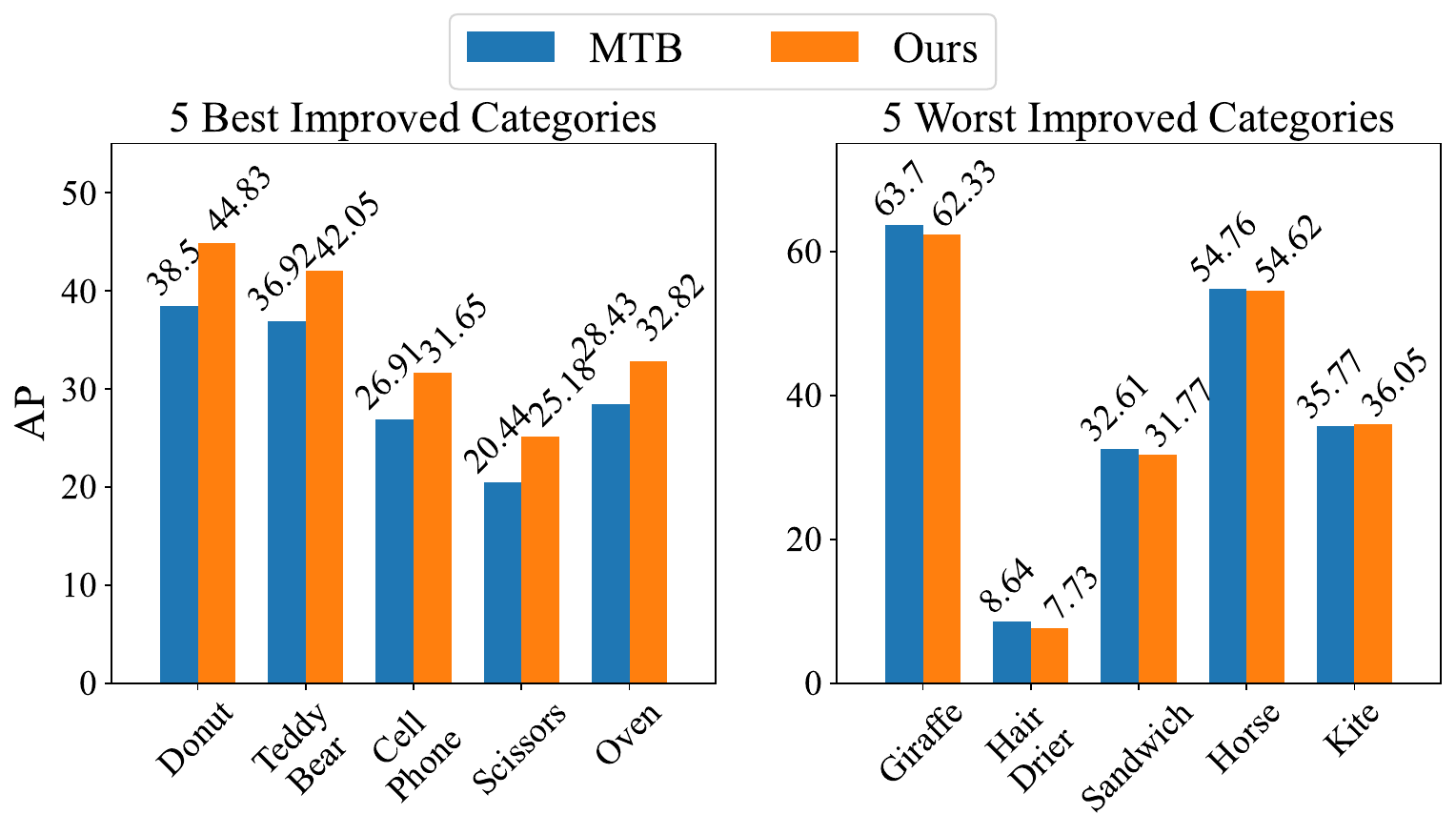}
    \caption{Comparison on AP results of the best improved categories (left) or worst improved categories (right) in COCO. }
    \label{fig:best5}
\end{figure}

We list the AP results of some categories from dataset COCO of our model and baseline ``MTB'', to see how our method performs in each category. The results for the 5 most frequent categories (object categories that have the most annotations) and least frequent categories are shown in Fig. \ref{fig:big5}.  The results for 5 best-improved categories (compared with baseline `MTB') and worst improved categories are shown in Fig. \ref{fig:best5}.  

It can be seen from the results that our model achieves a generally stable improvement in all the categories. For all the 80 categories on COCO, only 4 of them have a lower mAP value than baseline ``MTB''. Category ``giraffe'' suffers from a drop of 1.37\% in AP, which is the worst category of all. However, the best-improved category ``donut'' has seen an increase of 6.33\% in AP.  

\subsection{Ablation Study}
\label{ssec:Ablation}
To validate the effectiveness of each component of the proposed method, we carry out an ablation study on COCO and VOC in Table \ref{tb:ablation} \& \ref{a_tab:ablation_voc} respectively. 

The first row without any module added is the baseline ``MTB''. The results demonstrate that every component in our proposed method helps improve performance. Among them, AFI (Attention-based Feature Interactor) and  $L_\text{dis}$ play the most important parts. $L_\text{pse}$ also brings further improvement. On both datasets, AFI and $L_\text{fea}$ bring consistent improvements. Although $L_\text{dis}$ and $L_\text{pse}$ work on both datasets, improvement on VOC is quite limited. We found that prediction of FCOS on VOC are usually low in confidence, which easily brings noise to the supervision on the classification of the detection head, leading to the limited improvement.

\begin{table}[t]
\centering
\caption{Results of ablation study on COCO. 
}
\label{tb:ablation}
\begin{tabular}{cccc|ccc} 
\hline
AFI                       & $\mathcal{L}_\text{dis}$                    & $\mathcal{L}_\text{fea}$                    & $\mathcal{L}_\text{pse}$                    & COCO           & COCO\_A        & COCO\_B         \\ 
\hline
                          &                           &                           &                           & 36.18          & 50.79          & 31.63           \\
\checkmark &                           &                           &                           & 37.18          & 51.47          & 32.73           \\
                          & \checkmark &                           &                           & 37.12          & 51.89          & 32.52           \\
                          &                           & \checkmark &                           & 37.07          & 51.27          & 32.65           \\
                          &                           &                           & \checkmark & 36.88          & 51.04          & 32.47           \\
\checkmark & \checkmark & \checkmark &                           & 37.90          & 52.25          & 33.43           \\
\checkmark & \checkmark & \checkmark & \checkmark & \textbf{38.35} & \textbf{52.78} & \textbf{33.86}  \\
\hline
\end{tabular}

\end{table}

\begin{table}
\centering
\caption{Results of ablation study on VOC.}
\label{a_tab:ablation_voc}
\begin{tabular}{cccc|ccc} 
\hline
AFI       & $\mathcal{L}_\text{dis}$ & $\mathcal{L}_\text{fea}$ & $\mathcal{L}_\text{pse}$ & VOC   & VOC\_A & VOC\_B  \\ 
\hline
          &                                &                                &                                & 45.64 & 50.93  & 42.79   \\
\checkmark &                                &                                &                                & 46.70 & 52.11  & 43.78   \\
          & \checkmark                      &                                &                                & 45.78 & 50.84  & 43.06   \\
          &                                & \checkmark                      &                                & 46.90 & 53.14  & 43.54   \\
          &                                &                                & \checkmark                      & 46.22 & 51.74  & 43.24   \\
\checkmark & \checkmark                      & \checkmark                      &                                & 46.97 & 52.84  & 43.81   \\
\checkmark & \checkmark                      & \checkmark                      & \checkmark                      & \textbf{47.74} & \textbf{53.54}  & \textbf{44.62}   \\
\hline
\end{tabular}
\end{table}

\begin{figure*}

    \centering

    \includegraphics[width=0.98\linewidth]{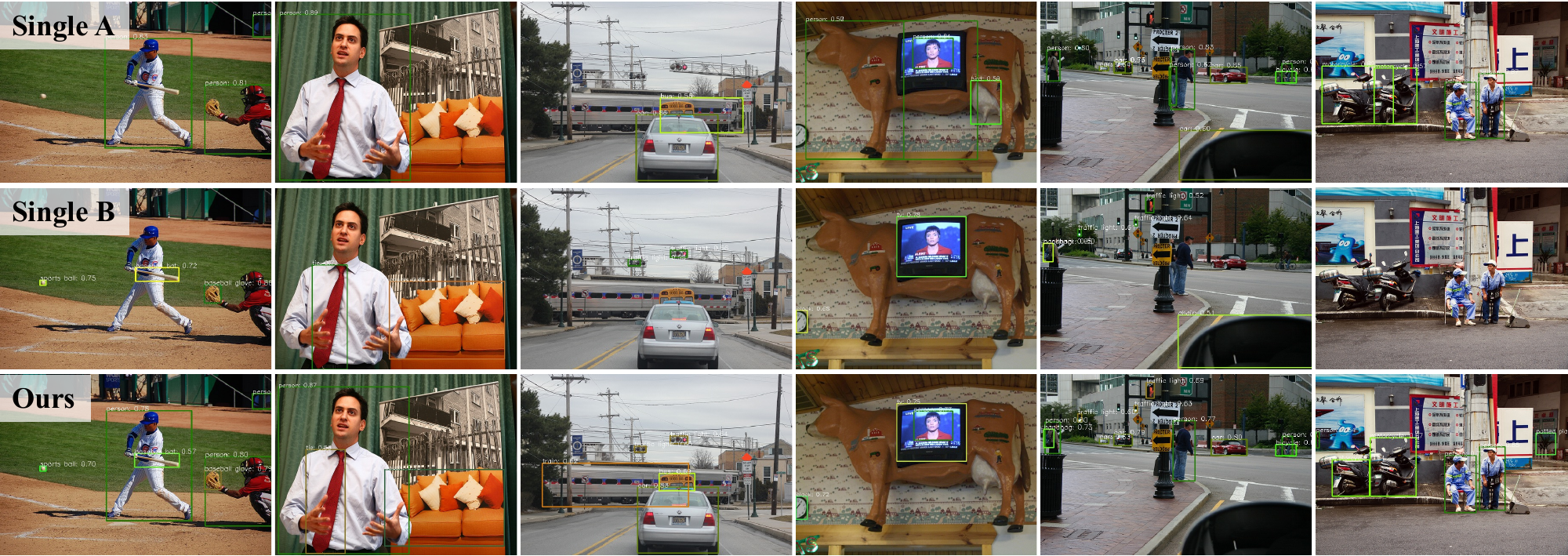}
    \caption{ 
    Visual results on COCO validation set. Our multi-dataset detection method (bottom row) has fewer missed objects or false detections than two single models for COCO\_A and COCO\_B separately (top two rows). Confidence threshold of visualization is set as 0.5.
    }
    \label{fig:teaser}
\end{figure*}

\subsection{Key Structure and Hyper-Parameters}
\label{ssec:exp_kd}

The proposed AFI module is one of the key structure in our model. Besides, there are two key hyper-parameters in KA strategy, and both of them are from $L_\text{pse}$. In this part, we will show how AFI module and these two hyper-parameters affect our model.

\subsubsection{AFI Module}

Experiments on COCO and VOC by replacing our proposed AFI module validates the superiority of the AFI module against several other modules such as ``CBAM'' \cite{woo2018cbam} and ``NDDR'' \cite{gao2019nddr} in multi-task learning.  The results are given in Table \ref{tb:afi}. ``CBAM(s)'' used the output of CBAM in a residual manner. Generally speaking, our AFI performs the best. It effectively combines features from different branches and adaptively lets features interact with attentions, thus improving detection on each branch. ``NDDR'' also utilizes cross-branch knowledge and thus performs better than other compared methods.

\begin{table}[]
\centering
\setlength{\tabcolsep}{4pt}
\caption{Replacing Attention-based Feature Interactor with other modules. The best result is denoted in bold. All the methods are based on our model, except that the feature interactor module is different. 
}
\label{tb:afi}
\begin{tabular}{c|ccc|ccc} 
\hline
Modules     & COCO  & COCO\_A & COCO\_B   & VOC            & VOC\_A         & VOC\_B           \\ 
\hline
CBAM    & 37.52 & 52.11 & 32.98 & 46.31          & 52.26          & 43.11           \\
CBAM(s) & 37.47 & 51.82 & 33.00 & 47.31          & 53.39          & 44.04           \\
NDDR    & 38.06 & 52.53 & 33.56 & 47.70          & 53.17          & \textbf{44.76}  \\
AFI (Ours)  & \textbf{38.35} & \textbf{52.78} & \textbf{33.86} & \textbf{47.74} & \textbf{53.54} & 44.62           \\
\hline
\end{tabular}
\end{table}

\subsubsection{Filter Threshold $\theta$}
The $\theta$ in Eq.(\ref{eq:pseudo_filter}) is a threshold to generate pseudo labels from images in other datasets. We vary the threshold from 0 to 1 with a step length 0.2. Results on COCO are given in Table \ref{tb:theta}. 
When $\theta=0$, all anchor points will be used for $L_\text{pse}$. It introduces lots of negative samples and performs bad. When $\theta=1$, $L_\text{pse}$ becomes a constant 0, which is the same case where no $L_\text{pse}$ is applied. Results show that proper thresholding matters. $\theta$ being too high results in few pseudo labels, while being too low brings too much noise.

\begin{table}[t]
\centering
\caption{Results with varying filter threshold $\theta$ in Eq.(\ref{eq:pseudo_filter}). 
}
\label{tb:theta}
\begin{tabular}{ccccccc} 
\hline
$\theta$   & 0     & 0.2   & 0.4   & 0.6   & 0.8   & 1      \\ 
\hline
COCO    & 38.19 & 38.10 & 38.35 & 38.05 & 38.39 & 37.90  \\
COCO\_A & 52.62 & 52.56 & 52.78 & 52.46 & 52.70 & 52.25  \\
COCO\_B & 33.69 & 33.59 & 33.86 & 33.56 & 33.93 & 33.43  \\
\hline
\end{tabular}
\end{table}

\begin{table}[t]
\centering
\caption{Results with varying the temperature $\hat{T}$ in Eq.(\ref{eq:pseudo_cls}). 
}
\label{tb:temperature}
\begin{tabular}{ccccccc}
\hline
$\hat{T}$       & 0.05  & 0.1   & 0.5   & 1     & 2     & 5     \\ \hline
COCO    & 37.92 & 38.35 & 38.25 & 38.60 & 38.17 & 38.23  \\
COCO\_A & 52.15 & 52.78 & 52.97 & 52.76 & 52.66 & 52.63  \\
COCO\_B & 33.49 & 33.86 & 33.66 & 34.19 & 33.66 & 33.75 \\
\hline
\end{tabular}

\end{table}

\subsubsection{Soft Pseudo Label Temperature $\hat{T}$ }

The $\hat{T}$ in Eq.(\ref{eq:pseudo_cls}) controls the flatness or sharpness of the softmax output distribution. Higher temperature will help the target net to learn more about the whole distribution of the teacher prediction and mine the relations between the most likely object category and the others. Lower temperature will make the softmax distribution sharper and provide supervision closer to one-hot code. We vary $\hat{T}$ from 0.005 to 5, and the results are given in Table \ref{tb:temperature}. Unlike in distillation, a relatively lower temperature produces better performance.

\subsection{Visual Results}
\label{ssec:Visual}

Detection results on some images from the validation set of COCO are shown in Fig. \ref{fig:teaser}. In general, it is easy to see that the ``Single'' models are only able to detect the classes of their own interest, while ours can locate all the objects. Besides, our method successfully decreases the number of missed objects or false positives. As is shown in the third column, the train is missed by ``Single'' model A but detected by ours. And for the fourth column, the cow is detected as a person by ``Single'' model A, while ours can avoid this mistake. A similar phenomenon can be observed in the other pictures.

\begin{figure}
    \centering
    \includegraphics[width=\linewidth]{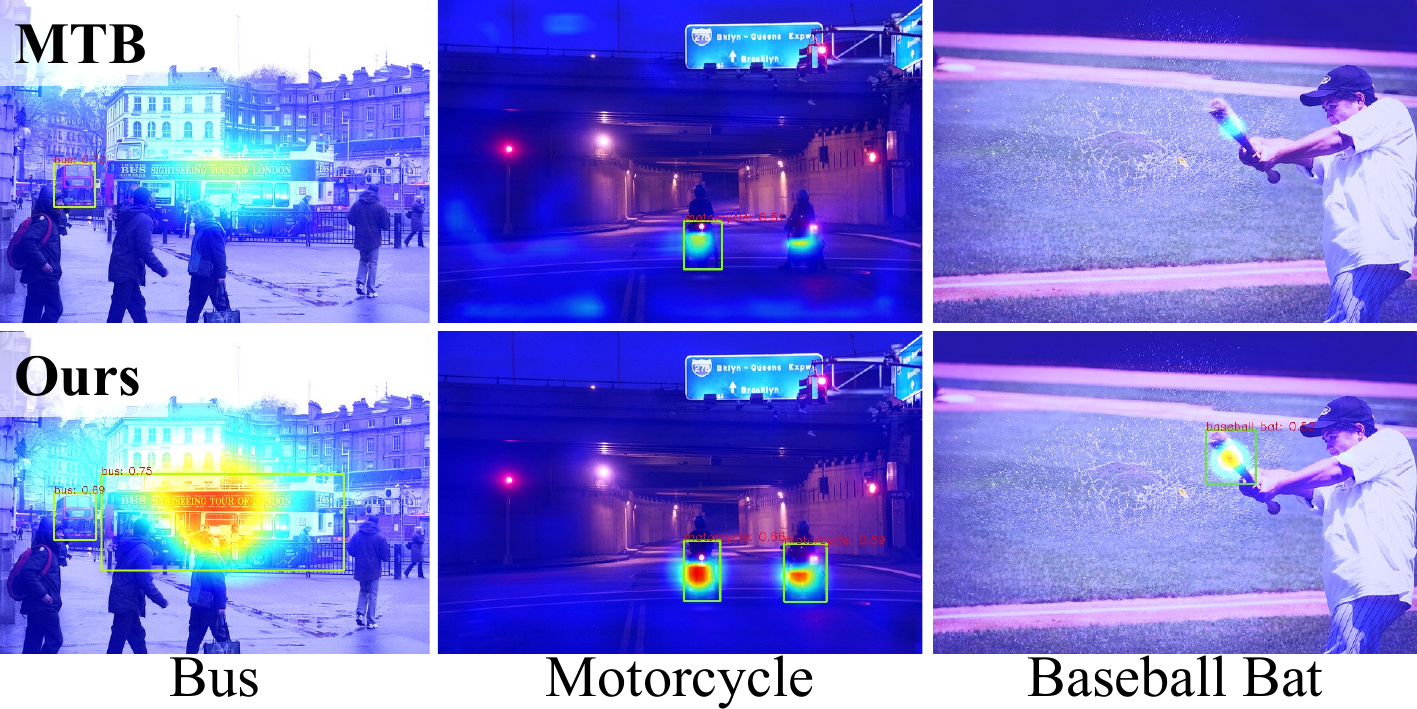}
    \caption{Visual results of confidence heatmaps of MTB (top row) and our method (bottom row) from COCO. Bounding boxes and heatmaps are shown only for the corresponding category. 
	Red indicates high confidence and blue means low.}
    \label{fig:heatmaps}
\end{figure}

The prediction confidence maps via \emph{heatmaps} are given in Fig. \ref{fig:heatmaps}. ``MTB'' misses many objects while our method is able to detect them all in these cases. Regardless of the size of these objects, our method effectively raises confidence in them.

\section{Experiments in extended scenarios}
Besides anno-incomplete multi-dataset detection, our proposed method could be applied to other multi-dataset detection scenarios where annotations are not strictly incomplete. Here, we apply our method in another two scenarios and present the results of these experiments.

\subsection{Multi Public Dataset Detection}

We tried to do a multi-dataset detection with COCO and VOC combined to see how our model performs when it is not strictly anno-incomplete. In fact, VOC categories are the subset of COCO categories. In this setting, the incomplete annotation will not be as serious as in Sec.\ref{ssec:OverallResults}.

For this experiment, we design one branch for COCO detection and another for VOC detection. Although there is a severe imbalance between dataset sizes, the weight for the total loss of each branch is still set to 1.0 and each batch still consists of an equal number of images from each dataset. Our model is trained with SGD for 180K iterations with the initial learning rate of 0.01 and a batch size of 16. The learning rate is reduced by a factor of 10 at 120K and 160K. Weight decay and momentum are set as 0.0001 and 0.9 respectively. The backbone is pre-trained on ImageNet \cite{imagenet}. We report the result at 130K iteration. Since there are great differences in the size of COCO and VOC (120k v.s. 16k), we fix the weight of loss from branch COCO as 1.5 and loss from branch VOC as 0.5. This method is denoted as ``$\text{Ours}_{\leftarrow \text{R101}}'$''.

The result is given in Table \ref{tb:overall_cocovoc}. Our method also shows competitive results. We implement the SOTA for semi-supervised detection ``UBT'' \cite{liu2021unbiased} for the same task too. The mAP for ``UBT'' on COCO is affected by 0.45\%, while ours by 0.36\%. However, we see a huge performance increase of 6.67\% on VOC, and ``UBT'' can only improve VOC mAP by 2.14\%. With the fixed loss weight for different branches, we are even able to get an increase of 1.51\% and 9.12\% in mAP on COCO and VOC respectively. Although both ``UBT'' and our model fully utilized other unlabelled data, our model implicitly mined the relations of categories across different datasets via an attention-based feature interactor, which helped our model to show a much stronger result.

\begin{table}[t]
\centering
\caption{ Results of mAP of our branch-interactive detector compared with other methods when trained with COCO and VOC in COCO evaluation protocol. The best result is denoted in bold.}
\label{tb:overall_cocovoc}
\begin{tabular}{c|cc|ccc} 
\hline
\multirow{2}{*}{Model}                 & \multicolumn{2}{c|}{COCO}          & \multicolumn{3}{c}{VOC}                                 \\ 
\cline{2-6}
                                       & mAP            & $\text{mAP}_{50}$ & mAP            & $\text{mAP}_{50}$ & $\text{mAP}_{75}$  \\ 
\hline
Single                                 & 38.63 & 57.41    & 47.97          & 74.98             & 52.15              \\
UBT                                    & 38.18          & 56.32             & 50.11          & 75.84             & 54.84              \\
$\text{Ours}_{\leftarrow \text{R50}}$  & 37.74          & 56.22             & 53.33          & 75.77             & 58.69              \\
$\text{Ours}_{\leftarrow \text{R101}}$ & 38.27          & 56.65             & 54.64 & 77.18    & 59.85     \\
$\text{Ours}_{\leftarrow \text{R101}}'$ & \textbf{40.14} &
\textbf{58.41}      & \textbf{57.09} & \textbf{79.33}    & \textbf{63.29}     \\
\hline
\end{tabular}
\end{table}

\subsection{Anno-exhaustive Multi-dataset Detection}
\label{ssec:anno-ex}

For ``Anno-incomplete Multi-dataset Detection'', there is a natural setting that constructed COCO\_A and COCO\_B have no common pictures and we manage to do that by erasing some of the annotations (as described in Sec.\ref{ssec:Dataset}).

One may wonder whether our method can achieve better results if we reserve all annotations when constructing the training set COCO(VOC)\_A/B. Such datasets are anno-exhaustive on all their images. Out of curiosity, we report results for this experiment named ``Ours (rsv)'' in this section. As we have all annotations reserved from the standard COCO and VOC dataset. We report another baseline ``Oracle'' here, which is an original FCOS detector trained with standard full annotations on COCO or VOC.

As is shown in Table \ref{tab:reserve}, with the same amount of annotations as ``Oracle'', ``Ours (rsv)''  is able to outperform ``Oracle'' by 2.18 in mAP on COCO. With a large proportion of annotations erased, ``Ours'' still achieves competitive results with ``Oracle''. 

The ``Oracle''s are detectors in conventional object detection scenarios and use all annotations, while ``Ours'' targets on an aforementioned new scenario, where a large proportion of the annotations are erased during dataset construction, thus it performs worse than ``Oracle''. In real-world applications where one has to train from multiple datasets, ``Oracle'' may not be obtained or compared. The results of ``Oracle'' are only as a reference of an $\emph{upper bound}$, to demonstrate the capability of our method of approaching the ideal virtual case even without the need for extra annotations.  

However, even if we can obtain all annotations in real-world applications on both datasets, ``Ours (rsv)'' is also able to outperform ``Oracle'' by a large margin, especially on COCO.

\begin{table}
\caption{mAP results on COCO\_A/B and VOC\_A/B when reserving all annotations during training set construction. The best result is denoted in bold. ``Ours (rsv)'' denotes our model trained when reserve all annotations during training set construction as described in Sec. \ref{ssec:anno-ex}.}
\label{tab:reserve}
\centering
\setlength{\tabcolsep}{2.5pt}
\begin{tabular}{c|ccc|ccc} 
\hline
               & COCO           & COCO\_A        & COCO\_B        & VOC            & VOC\_A         & VOC\_B          \\ 
\hline
MTB            & 36.18          & 50.79          & 31.63          & 45.64          & 50.93          & 42.79           \\
Ours           & 38.35          & 52.78          & 33.86          & 47.74 & 53.54  & 44.62           \\
Ours (rsv) & \textbf{40.81} & \textbf{54.01} & \textbf{36.69} & \textbf{48.17} & \textbf{53.56} & \textbf{45.27}  \\
Oracle         & 38.63          & 52.20           & 34.48          & 47.97          & 52.56          & 45.09           \\
\hline
\end{tabular}
\end{table}

\section{Conclusion}
\label{sec:Conclusion}
In this paper, we first proposed a novel and important problem of ``Anno-incomplete Multi-dataset Detection''. It aims at detecting all objects of interest based on multiple partially annotated datasets, where only the objects of its own interest are annotated for each dataset and those concerned in other datasets are not. This problem has hardly been studied, but is of great significance in real world applications.

To tackle this problem, we proposed a branch-interactive multi-task detector with an attention-based feature interactor, which helps to mine the relations implicitly among categories from different datasets.
Besides, we proposed a amalgamation training strategy to handle cross-domain variations from different branches, and use pseudo labels to explicitly exploit the unlabeled objects.
Experiments on COCO and VOC validated the effectiveness of the proposed method.  An improvement of 2.17\% and 2.10\% respectively in mAP can be achieved. 
Although only two-dataset detections were exhibited in experiments, our method can be easily extended on 3 or more datasets.



{\appendices
\section{Training Details of Our Proposed Method and Baselines}
\label{a_sec:training}
\subsection{Our Proposed Model}
For COCO, Our model is trained with SGD for 90K iterations with the initial learning rate of 0.01 and a batchsize of 16. The learning rate is reduced by a factor of 10 at 60K and 80K. Weight decay and momentum are set as 0.0001 and 0.9 respectively. The backbone is pretrained on ImageNet \cite{imagenet}. We report the best result around 70K. And for VOC, all the other parameters are the same except that the total iteration is 9450 and learning rate is reduced at 6300 and 8400. We report the result at the last checkpoint.

Specially, we follow the settings in \cite{zhou2021simple} and make sure that each batch consists of equal number of images from each dataset. For amalgamation training strategy, teachers of Res101 or Res50 are adopted, which are trained with the same setting for single models. 
$\omega_\text{det}$, $\omega_\text{dis}$, $\omega_\text{fea}$ and $\omega_\text{pse}$ are all simply set to 1.0 when training on COCO. When training on VOC, $\omega_\text{dis}$ is set to 0.1 while others remain as 1.0.  Distillation temperature $T$, soft pseudo label temperature $\hat{T}$ and filter threshold $\theta$  are set to 5.0, 0.1 and 0.4 respectively. 

We implement our methods with PyTorch 1.4.0 and experiments are carried on 8 GPUs of NVIDIA V100.

\subsection{Baselines}

\subsubsection{MTB, UODB}
``MTB'' is the base model of ours, where there is no AFI module or KA strategy. ``UODB'' \cite{wang2019towards} is implemented with several domain attention modules within the backbone of ``MTB''. These two models are all trained with the same strategy as our proposed model.

\subsubsection{Single}
For COCO\_A/B, single models are trained with SGD for 45K iterations with the initial learning rate of 0.01 and a batchsize of 16. The learning rate is reduced by a factor of 10 at 30K and 40K. Weight decay and momentum are set as 0.0001 and 0.9 respectively. For VOC\_A/B, the total iteration is 4725 and learning rate is reduced at 3150 and 4200.

\subsubsection{Oracle}
The oracle model is a single model trained on the standard COCO or VOC. For COCO, the model is trained with SGD for 90K iterations with the initial learning rate of 0.01 and a batchsize of 16. The learning rate is reduced by a factor of 10 at 60K and 80K. For VOC, the total iteration is 9450 and learning rate is reduced at 6300 and 8400.

\subsubsection{Brutal}
``Brutal'' model is trained on the merged dataset of partitions on COCO (e.g. directly concatenated dataset from COCO\_A and COCO\_B). Compared to the standard COCO, the total amount of images remains the same except that there are some annotations erased. Training settings are the same as ``Oracle''.

\subsubsection{SMD}
``SMD'' \cite{zhou2021simple} is to learn a unified label space from COCO\_A/B. However, we ran the open-source code of SMD and found there is no label to merge. That's to say, the merged dataset is the same as ``Brutal''. Training settings are the same as ``Oracle''. And the final results are almost the same to ``Brutal''. 

\subsubsection{Pseudo}
\label{a_sec:setting_pseudo}
The baseline ``Pseudo'' is to use the ``Single'' models to predict  boxes  on  other  datasets  (single  model  for  COCO\_A to  infer  on  COCO\_B).  We  take  the  predictions  as  annotations  and  merge  all  the  datasets  along  with  the  predictions into one dataset, and finally train a single-branch detector. 
When converting predictions into annotations, NMS of threshold 0.2 is applied to filter highly overlapped boxes. Besides, score threshold 0.6 is adopted to keep the highly confident predicted boxes. For VOC, the score threshold is set to 0.2 however. As predictions on VOC have a generally low confidence score, higher threshold will only incorporate too few or no boxes.
The model is trained on the merged dataset with the same training parameters as ``Oracle''.

\section{Analysis on the Performance of Baseline ``Brutal'' and ``Pseudo''}
``Brutal'' is trained on the directly combined dataset of COCO\_A/B as described in Sec. \ref{ssec:Baselines}. 

Compared with ``Single'' on COCO\_A, the dataset got bigger, but it also introduced dirty unlabeled COCO\_A objects from images of COCO\_B, which were taken as the negative ``background'' and were misleading. It would confuse the detector and result in drop in accuracies. Compared with ``MTB'', ``Brutal'' has the same amount of annotations and images. But ``MTB'' handles COCO\_A on its own specific branch, avoiding misleading information from unlabeled COCO\_A objects from images of COCO\_B. Thus, ``MTB'' shows better results than ``Brutal''.

As is shown in Table \ref{tb:overall}\&\ref{tb:overall_voc}, it is easy to find that ``Pseudo'' behaves well on COCO but poor on VOC.

We assume the difference comes from the datasets. 71 classes in COCO have more than 2000 annotations, while only 5 in VOC. 67.5\% of COCO classes have over 5000 annotations, while it is only 5\% in VOC. Smaller dataset makes it harder to learn for FCOS detector, leading to the low confidence in VOC predictions. We tried unified thresholds on COCO and it is easy to find one that will generate good performance. But when we tried different unified thresholds for ``Pseudo'' on VOC, none of them gets better. A high threshold will leave no predictions left while lower threshold will introduce too much noise.
However, if we carefully tune different thresholds for each class on VOC, the results get way better (see ``UBT'' in Table \ref{tb:overall}\&\ref{tb:overall_voc}). To keep the proposed architecture simpler and easier to use, we chose to use unified threshold for all classes, instead of maximizing accuracy. However, adaptations are free in applications.  
Inconsistency of performance on COCO and VOC is also found in the ablation study of $L_\text{pse}$ and $L_\text{dis}$ in Sec.\ref{ssec:Ablation}, which we believe comes from the similar reason.

\section{Comparison with KDOD and CIOD}
\label{asec:compare}
In this section, we analyze and compare ``Anno-incomplete Multi-dataset Detection'' with \emph{Knowledge Distillation Object Detection} (KDOD) and \emph{Class Incremental Object Detection} (CIOD).

KDOD aims at getting a compressed model on a single dataset while we learn a versatile detector from multiple datasets, which is very different from KDOD. The most related SOTA in KDOD is Unbiased Teacher (UBT)\cite{liu2021unbiased}
which utilizes knowledge distillation to train a student model in a semi-supervised way. As shown in Table \ref{tb:overall}, UBT achieves an mAP of 37.00 on COCO while ours achieves 38.31.

CIOD aims at training a model to detect extra classes without re-training on the original dataset, while we train on multiple off-the-shelf datasets (possibly with similar sizes, or not) simultaneously and also utilize the correlations among the datasets, which is a more convenient, thorough and once-and-for-all manner in our new scenario.

\begin{table}
\centering
\caption{mAP50 of CIOD results on VOC from ORE\cite{joseph2021open}. All models are based on Faster-RCNN.  }
\label{a_tab:ore}
\setlength{\tabcolsep}{4pt}
\begin{tabular}{lrrr} 
\hline
Model       & VOC (First 10) & VOC (New 10) & VOC (All 20)  \\ 
\hline
All 20      & 70.82          & 70.20        & 70.51         \\
First 10    & 71.18          & 0            & 35.59         \\
New 10      & 2.26           & 70.35        & 36.31         \\
Faster ILOD & 69.76          & 54.47        & 62.16         \\
ORE         & 60.37          & 68.79        & 64.58         \\
\hline
\end{tabular}

\end{table}

We cite Table 3 from the recent SOTA for class incremental object detection - ORE \cite{joseph2021open} here as Table \ref{a_tab:ore}. 
``All 20'' and ``First 10'' are a single Faster-RCNN trained on all 20 classes and first 10 classes of VOC respectively. ``New 10'' is firstly trained on the first 10 classes and then trained on the last 10 classes of VOC. ``Faster ILOD'' \cite{PENG2020109} is another SOTA for CIOD. 
ORE still performs worse than ``All 20'' which has access to annotations from both sets of classes at any stage. Note that
the ``All 20'' model setting is exactly the same as ``Brutal'' in our paper, and is outperformed by our proposed method by a large margin. Therefore, it is unfair for most CIOD methods to be compared with us as they target on a different task and also has lower performance. So we decided not to compare with them in the main script.

}

 
%
 
\bibliographystyle{IEEEtran}
\bibliography{ref}

\end{document}